\documentclass[sigplan,screen,anonymous=false]{acmart}
\AtBeginDocument{%
  \providecommand\BibTeX{{%
    \normalfont B\kern-0.5em{\scshape i\kern-0.25em b}\kern-0.8em\TeX}}}

\setcopyright{acmcopyright}
\copyrightyear{2023}
\acmYear{2023}
\acmDOI{XXXXXXX.XXXXXXX}

\acmConference[E-Energy '23]{On the utility and fairness of pre-trained models in modeling distributed energy resources}{June 20--23,
  2023}{Orlando, FL}
\acmBooktitle{E-Energy '23: 14th ACM International Conference on Future Energy Systems,
 June 20--23, 2023, Orlando, FL} 
\acmPrice{15.00}
\acmISBN{978-1-4503-XXXX-X/18/06}

\begin{document}

\title{On the contribution of pre-trained models to accuracy and utility in modeling distributed energy resources}

\author{Hussain Kazmi}
\email{hussainsyed.kazmi@kuleuven.be}
\affiliation{%
  \institution{KU Leuven}
  \country{Belgium}
}

\author{Pierre Pinson}
\affiliation{%
  \institution{Imperial College London}
  \country{UK}}







\renewcommand{\shortauthors}{Kazmi and Pinson}

\begin{abstract}

Despite their growing popularity, data-driven models of real-world dynamical systems require lots of data. However, due to sensing limitations as well as privacy concerns, this data is not always available, especially in domains such as energy. Pre-trained models using data gathered in similar contexts have shown enormous potential in addressing these concerns: they can improve predictive accuracy at a much lower observational data expense. Theoretically, due to the risk posed by negative transfer, this improvement is however neither uniform for all agents nor is it guaranteed. In this paper, using data from several distributed energy resources, we investigate and report preliminary findings on several key questions in this regard. First, we evaluate the improvement in predictive accuracy due to pre-trained models, both with and without fine-tuning. Subsequently, 
we consider the question of fairness: do pre-trained models create equal improvements for heterogeneous agents, and how does this translate to downstream utility? Answering these questions can help enable improvements in the creation, fine-tuning, and adoption of such pre-trained models.

\end{abstract}






\maketitle





\section{Introduction}
Data-driven modelling of complex dynamical systems typically requires access to large amounts of data to generalize well \cite{han2021pre}. In the absence of large datasets, model performance tends to be sub-par in regions of state-space that have been poorly explored \cite{chung2019slice}. Additionally, models trained with limited data may learn relationships that are not causal \cite{lapuschkin2019unmasking}. Using such models for control can therefore lead to undesirable outcomes in high stakes decision-making \cite{rudin2022interpretable}, such as in the operation of critical infrastructure or decisions that directly affect human well-being.

However, in many domains, outside of computer vision and natural language processing, collecting large amounts of data is often infeasible \cite{kazmi2021towards}. This is frequently the case with predicting demand for new products or services, where data-driven models are not feasible due to the cold-start problem \cite{moon2020solving}. Another case where observational data tends to be limited is due to expensive sensing or communication infrastructure, as well as privacy concerns \cite{mcdaniel2009security, morel2022your}. This is frequently the case in the energy domain, where data on both demand and generation side is typically gathered in the form of time series, which need to be modelled and predicted. However, installing and maintaining sensors to continuously monitor electricity demand and generation at high spatiotemporal resolutions is an expensive and often privacy-compromising operation.

As a result, several solutions have been proposed in recent years to reduce the data requirements or sample complexity of learning algorithms, including via the use of pre-trained models (PTMs), which rely on simulated or historically observed data, thereby eschewing the need for observational data \cite{han2021pre}. More recently, this has also seen significant interest in the modeling and optimization of distributed energy resources \cite{peirelinck2022transfer, didden2021sample}. In this paper, we dive deeper into the topic to consider how (or whether) PTMs can accelerate the data-driven modelling of distributed energy resources, e.g. hot water storage systems. More concretely, we attempt to answer the following questions: (1) do PTMs outperform purely data-driven dynamics models; (2) (how) does the size of the source and target corpus, and fine-tuning affect the PTM's performance; and (3) do potential improvements induced by PTMs create uniform utility for heterogeneous agents. The remainder of this paper is organized as follows. In section 2, we describe the methodology we follow subsequently. Section 3 presents key results, while Section 4 concludes the paper with an eye to future research directions.


\section{Methodology}

\subsection{Data and setup}

To answer the afore-mentioned questions, we utilize detailed (quarter-hourly) data spanning over one year from several identical heat pump hot water storage systems in a net-zero energy neighborhood, constituting a corpus of 24 buildings. 
The general idea is to construct a model for state estimation of the hot water storage system in a way that can then used for downstream control (e.g. to reduce energy demand, maximize user comfort or minimize grid load etc.).
Of the entire corpus, 8 systems were partitioned into a separate source set, $S$, for which data was assumed to be available before the start of the experiment, i.e. this data can be used off-the-shelf to create PTMs. The remaining 16 buildings are marked as the target set, $T$, which can be used to construct 'local' models for the hot water systems. A local model identifies that only observational data from a particular building is used during the modeling process. However, for each individual building, even after an entire year of acquisition, this dataset is rather small at a few thousand training examples. Even ignoring the effects of serial and autocorrelation on the information content, this corpus is smaller than even a single high-resolution image. 

Even though the systems are identical, the users and operating conditions are not. Consequently, each system generates data which is distributed differently, due to the sampling bias these two factors induce. This means the datasets observed at different systems, and used to train (and evaluate) the local models can be significantly different.


\begin{figure}
    \centering
    \includegraphics[width=.65\linewidth]{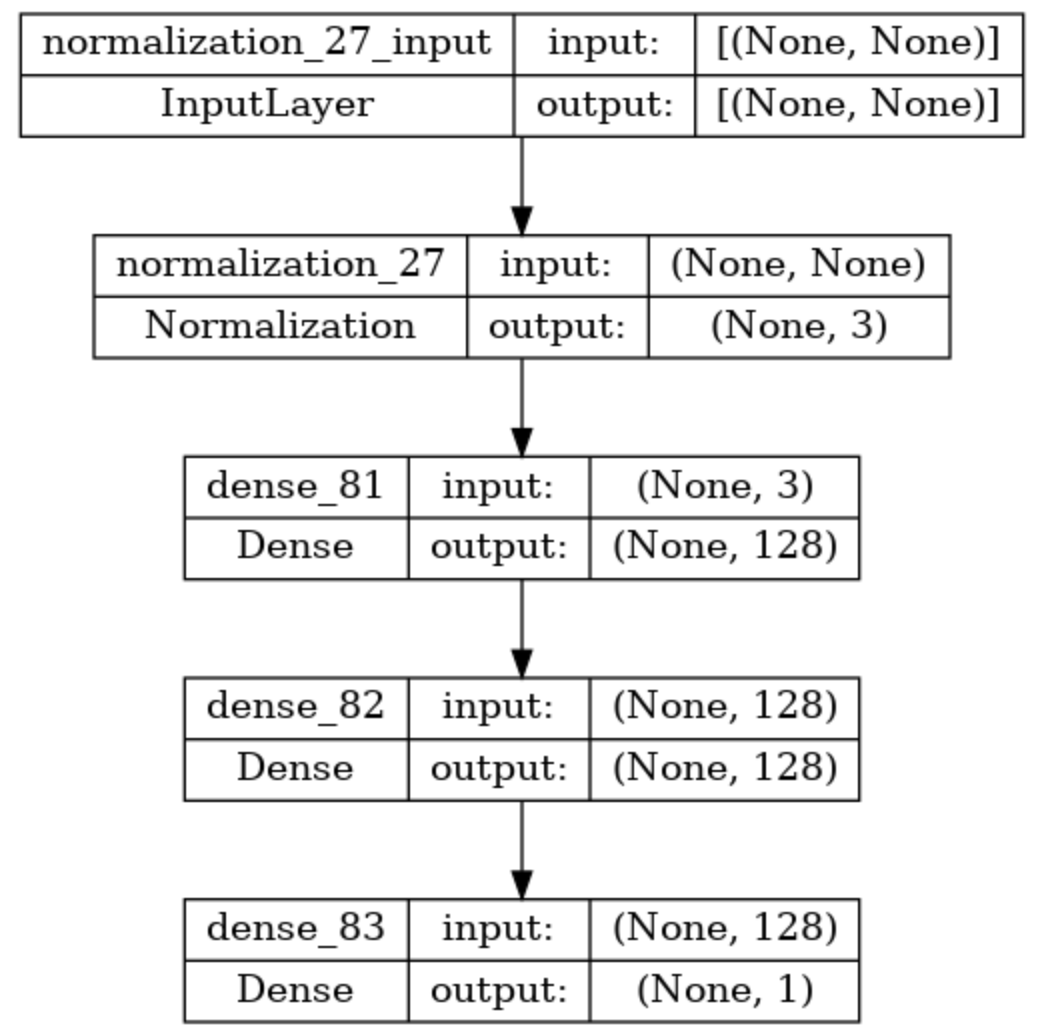}
    \caption{Neural architecture for the trained models}
    \label{fig:arch}
\end{figure}

\subsection{Local models}

We pose the problem as a standard one-step ahead regression problem, $y = f(x)$. Here, the model, $f(.)$, predicts the temperature at the mid-point, $T_m$, in the storage vessel. To do so, it uses three fields as input: (1) the time since the last reheat cycle, $t$, (2) the hot water demand since the last reheat cycle, $w$  and (3) the initial temperature of the vessel at its mid-point, $T_0$. These features are extracted from three observational time series respectively: user hot water demand, heat pump operation (i.e. active or idle), and the observed temperature at the mid-point in the storage vessel. 
This means that at any given time point, the entire history of the three time series can be used to construct the input feature matrix, which is then ingested by the function approximation model. Note that, by virtue of including time since the last reheat cycle in the input features, arbitrary future time horizon predictions can be obtained.

The model, $f(.)$ is chosen to be a neural network with two fully connected hidden layers and a normalization layer (shown in Fig. \ref{fig:arch}). It utilizes the ReLU activation function and L2 regularization, and is fit using the mean absolute error (MAE) loss function (i.e. it minimizes the error when predicting the mid-point temperature in the vessel). The reason for choosing a neural network as the function approximation technique is motivated by its prevalence in PTM literature. While theoretically linear models can be used in a similar fashion, they are incapable of modelling complex functions and non-linear dynamics, as is required here (heat losses are inherently non-linear).

The sampling bias in the observational data manifests as a distributional shift in the three features for different households, and predominantly affects the local models, which are trained only using data collected at each site. These local models are trained at four distinct time points, namely after 4, 8, 16, and 32 weeks. The holdout test dataset for each site is derived locally by sampling the last three months of observations, and consequently does not overlap with the training set. As such, it is still subject to many of the same sampling biases (e.g., a high accuracy score for a local model on a local test set does not guarantee broad generalization). 

\subsection{Pre-trained models}

Instead of relying solely on observational data to fit model parameters, PTMs utilize what is essentially a sequential two-stage process. First, a model is trained on (abundant) simulated or prior observational data. Second, it is fine-tuned using (sparse) observational local data. 
The pre-training relies primarily on the fact that the processes being modelled are identical or, at the very least, share some similarities. In addition to similarity, the amount of the pre-collected training data (i.e. source corpus size) is also important. Where both the quantity and quality of pre-collected data is sufficient, PTMs can lead to large speed-ups in the modelling of dynamical systems (typically observed as better initial accuracy, learning rate, and asymptotic accuracy). 
To make the comparison fair, we employ the same model architecture as before for creating the initial PTM, and investigate the effect of several variations, including: (1) the source corpus size (i.e. how much data was available to train the PTM), (2) fine-tuning (i.e. whether the PTM employs fine-tuning using local data or not), and (3) the target corpus size (i.e. the local dataset size, which is only relevant when we utilize fine-tuning).



More concretely, to investigate the effect of the source corpus size, we created two models. The first model used a large training corpus, including data from eight systems (these systems were not included in the local models discussed above) for eight months. The second model utilized a small corpus, referring to data from only a single system for a duration of eight months. After training to convergence, these two PTMs were then deployed in the households, where their performance was evaluated using the same local holdout datasets and methodology as described previously. 

In the case where these two PTMs were not fine-tuned using local data, their performance on the hold-out test set for any given system did not change over time (the holdout set is kept fixed to avoid any reporting bias).
On the contrary, to investigate the impact of fine-tuning, given varying amounts of target data, 
we followed the same logic as that of local models. Both the small and large corpus PTMs were fine-tuned at four distinct points in time: i.e. at 4, 8, 16, and 32 weeks. The fine-tuning process is carried out at a lower learning rate and for fewer iterations, when compared with the training of the PTMs or local models. The models are evaluated using the same method as that described previously.


\section{Results}

In this section, we present the most important results comparing local models against different variants of PTMs. Subsequently, we take a look at the implications this has on donwstream operations.

\subsection{The importance of corpus size and fine-tuning}

\begin{figure}
    \centering
    \includegraphics[width=.8\linewidth]{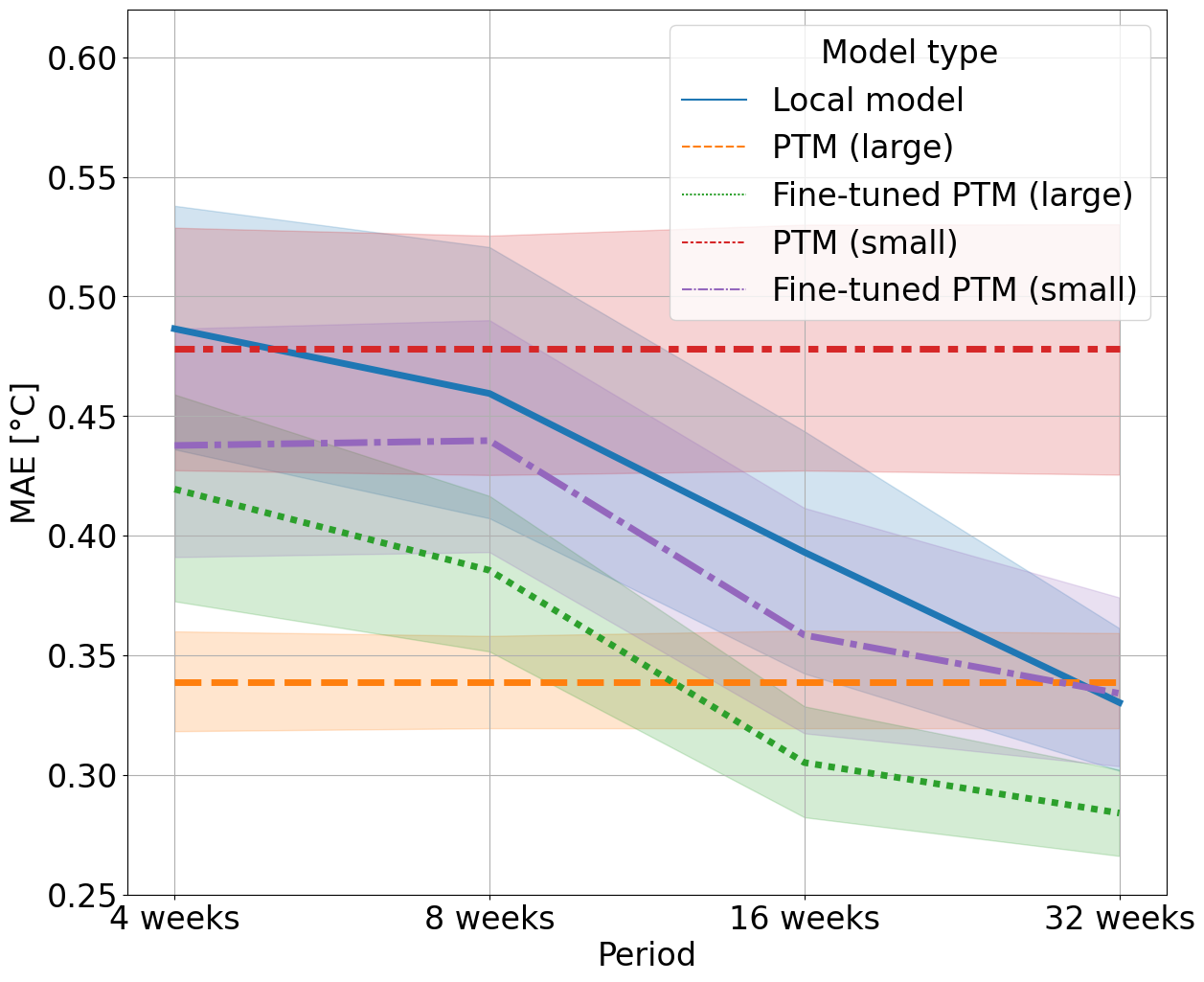}
    \caption{Mean absolute error (MAE) on holdout test dataset for different model configurations with increasing amounts of training data, averaged over 16 systems}
    \label{fig:local-PTM-FT}
\end{figure}

Fig. \ref{fig:local-PTM-FT} shows the predictive performance of the different model configurations, using training corpus size (or observation period) as the x-axis. As expected, increasing amounts of data, either during the pre-training phase or the local acquisition phase, improve performance. This holds for both local models, as well as small and large fine-tuned PTMs. Asymptotically, the small training corpus PTM performs the worst, but is still comparable to a local model during the initial phase, when the local models have seen only four weeks of on-site training data. Increasing amounts of observational data lead to improvements in the local models, with the average error falling from around 0.5°C after 4 weeks to 0.33°C after 32 weeks. Fine-tuning both PTMs shows considerable potential to improve this performance further, especially when only limited amounts of locally observed data has been collected. Even so, the local and fine-tuned small corpus PTM perform at roughly the same accuracy level as the large corpus PTM without any fine-tuning. With fine-tuning, the large corpus PTM significantly outperforms the local models with the average error falling as low as 0.28°C after 32 weeks.

A curious trend emerges here. While the PTM with fine-tuning ends up being the best performing model at the end of 32 weeks, the best model in the low observational data regime is actually the large PTM without any fine-tuning. This makes sense since the PTM has arguably already learnt the correct system dynamics, and only a small amount of training data does not allow it to further reduce its error on the (biased) evaluation test dataset for each specific system. This also shows that perhaps the fine-tuning might be too aggressive in the low data regime. 

\subsection{The importance of choosing the correct PTM}

\begin{figure*}
    \centering
    \includegraphics[width=.785\linewidth]{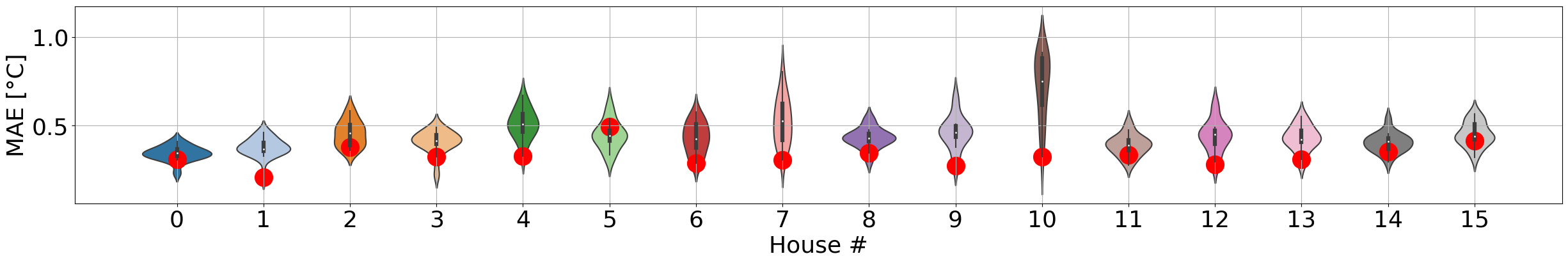}
    \caption{The cross-performance of local models evaluated on holdout test set for each individual building; the large markers represent the performance of the model trained using data from the system}
    \label{fig:cross-PTM}
\end{figure*}

The small corpus PTM (without fine-tuning) exhibits significant deviation in the improvements it can bring to different systems (note the wide uncertainty bands plotted in Fig. \ref{fig:local-PTM-FT}). In some instances, the performance of these models can even begin to rival that of much better performing models (e.g. the large corpus PTM with fine-tuning). To investigate this effect further, we treat each individual local model 
trained for 8 months as a small corpus PTM. This leads to 16 additional candidate small source corpus PTMs. As there is no overlap in training or evaluation data, this does not pose any data leakage concerns. Fig. \ref{fig:cross-PTM} shows results for evaluating these models without fine-tuning, and presents two key insights. The first is the enormous amount of diversity in accuracy for most systems when data from a different system is used to learn their dynamics. This is all the more surprising because the systems are identical, and the only difference is the sampling bias due to occupant and controller behavior. Second, we see that the model trained on the system's data itself is usually, but not always the best performing model. This seems counter-intuitive at first but there are several explanations for it, including (1) non-stationarity in the training and evaluation data, i.e. the local dataset underwent some distributional shifts, and (2) data quality, i.e. the data (training and/or validation) was rather noisy to begin with.

\section{Conclusions and future work}

Fig. \ref{fig:cross-PTM} has already demonstrated that there is a large difference in the use of one source model vs. a different one, even when they are sourced from similar data sources. This has profound implications in terms of downstream utility. In our downstream experiments, we found that this caused significant differences while optimizing heating demand for the storage vessel, constrained on occupant comfort. Local models were, for most houses (14 out of 16), unusable after the initial 4 week period, to the extent that they led to control policies that significantly violated user comfort due to not reheating the storage vessel at all or at incorrect times. After 32 weeks, the situation was reversed, as most local models (12 out of 16) were able to control the vessel in a near optimal manner. Surprisingly enough, the large corpus PTM, with no fine-tuning, led to the optimal control strategy for all systems. The small corpus PTM performed roughly on par with the local model with only 4 weeks of data (i.e. it was largely unusable).

This obviously begs several question. First, if a large source corpus model with no fine-tuning can be used to control distributed energy resources, then local data can be used solely for validation (or fine-tuning). This means local data can stay on the user's premises, leading to a simultaneous improvement in both downstream utility and reduction in privacy leakage. This would not be possible otherwise with classical machine learning based solutions. Furthermore, where large pre-trained models are not available, Fig. \ref{fig:cross-PTM} shows that it is possible to use a carefully chosen small source corpus PTM to achieve comparable, if not better, performance as using a local model. In this paper, we have not explored in greater depth the drivers for this heterogeniety although similarities in feature space certainly seem to play a role here. This is an important avenue for future research.

The results presented in the paper also have important implications for future electricity and data markets design. While scarce at the moment, movements to collect more behind-the-meter data are gathering momentum, raising several privacy concerns. By utilizing PTMs, this data can stay on-premises, but valuing their contribution in monetary terms, especially tying in with downstream utility, is an open and important direction for future research.






\begin{acks}

Hussain Kazmi acknowledges support from FWO, Belgium in the preparation of this manuscript.

\end{acks}

\bibliographystyle{unsrt}
\bibliography{sample-base}

\appendix

\end{document}